\documentclass{article}

\PassOptionsToPackage{square, numbers}{natbib}
\usepackage[final]{nips_2017}
\usepackage{citrine}

\usepackage[utf8]{inputenc} 
\usepackage[T1]{fontenc}    
\usepackage{hyperref}       
\usepackage{url}            
\usepackage{booktabs}       
\usepackage{amsfonts}       
\usepackage{nicefrac}       
\usepackage{microtype}      

\usepackage{subcaption}
\usepackage{verbatim}

\title{Overcoming data scarcity with transfer learning}

\author{
  Maxwell L.~Hutchinson\\
  Citrine Informatics\\
  Redwood City, CA 94063\\
  \texttt{maxhutch@citrine.io}\\
  \And
  Erin Antono \\
  Citrine Informatics\\
   Redwood City, CA 94063\\
  \texttt{eantono@citrine.io} \\
  \And
  Brenna M. Gibbons \\
  Stanford University \\
  Stanford, CA 94305 \\
  \texttt{brennamg@stanford.edu}
  \And
  Sean Paradiso \\
  Citrine Informatics\\
  Redwood City, CA 94063\\
  \texttt{sparadiso@citrine.io} \\
  \And
  Julia Ling \\
  Citrine Informatics\\
  Redwood City, CA 94063\\
  \texttt{jling@citrine.io} \\
  \And
  Bryce Meredig \\
  Citrine Informatics\\
  Redwood City, CA 94063\\
  \texttt{bmeredig@citrine.io} \\
}

\begin{document}

\maketitle

\begin{abstract}
Despite increasing focus on data publication and discovery in materials science and related fields, the global view of materials data is highly sparse.
This sparsity encourages training models on the union of multiple datasets, but simple unions can prove problematic as (ostensibly) equivalent properties may be measured or computed differently depending on the data source.
These hidden contextual differences introduce irreducible errors into analyses, fundamentally limiting their accuracy.
Transfer learning, where information from one dataset is used to inform a model on another, can be an effective tool for bridging sparse data while preserving the contextual differences in the underlying measurements.
Here, we describe and compare three techniques for transfer learning: multi-task, difference, and explicit latent variable architectures.
We show that difference architectures are most accurate in the multi-fidelity case of mixed DFT and experimental band gaps, while multi-task most improves classification performance of color with band gaps.
For activation energies of steps in NO reduction, the explicit latent variable method is not only the most accurate, but also enjoys cancellation of errors in functions that depend on multiple tasks.
These results motivate the publication of high quality materials datasets that encode transferable information, independent of industrial or academic interest in the particular labels, and encourage further development and application of transfer learning methods to materials informatics problems.

\end{abstract}

\section{Introduction}

Data-driven methods have demonstrated the potential to fundamentally change how new materials are discovered and optimized.
Machine learning-based models have guided the laboratory discovery of novel nickel-based superalloys for aircraft engines~\cite{conduit2017design}, thermoelectric materials for waste heat capture~\cite{gaultois2016perspective}, organic light-emitting diodes~\cite{gomez2016design}, Heusler-structured compounds~\cite{oliynyk2016high}, and shape-memory alloys with potential actuator applications~\cite{xue2016accelerated}.
While long-established machine learning algorithms such as support-vector regression~\cite{lee2016prediction} and random forest~\cite{ward2016general} have been widely employed in the materials informatics community for some time, more recent methods such as convolutional neural nets, variational autoencoders, recurrent neural nets, and generative adversarial nets have recently gained adoption in the materials informatics community for tasks including microstructure classification~\cite{decost2017exploring}, crystal structure classification~\cite{park2017classification}, and molecular structure generation~\cite{gomez2016automatic,guimaraes2017objective,yang2017chemts}.

The lack of sufficient training data is a major challenge in the application of machine learning to materials science, which, as a field, suffers from an unusually high data acquisition cost compared to other domains in which machine learning has proven useful.
Unsurprisingly, the largest materials-related training sets--which typically consist of $\sim10^5$ examples--are derived from physics-based simulations, as opposed to the results of laboratory experiments.
While these data encode rich physical insight, they also introduce systematic approximation errors due to closure approximations, limitations in the applied theories, or simply through the discretization of time and space.
Without ground-truth experimental values to learn corrections, machine learning methods based on computational values are fundamentally limited in their ability to solve real world engineering problems. 
It is perhaps the removal of this limitation that most urgently motivates the use of transfer learning (TL) methods in materials science.

Transfer learning (TL) is a class of methods that transfer information between distinct learning tasks.
In this way, TL can leverage a larger dataset generated through a less expensive means (\eg computation) to improve models for a more limited dataset generated via a more expensive means or pool information from several related but disjoint datasets.
TL can also couple predictions, better preserving correlations between properties and/or benefiting from cancellation of error.

In this paper, we explore the effectiveness of explicit latent variable, multi-task, and difference architectures in the context of two materials engineering problems: i) predicting the (experimental) band gap and color of solid-state crystalline compounds and ii) predicting activation energies for three important reaction steps in the reduction of nitric oxide (NO) and subsequently identifying the rate determining step.
\sref{methods} introduces the three TL architectures and describes validation methods.
\sref{bandgap} presents results for transferring from computational band gap data to experimental band gap and color data.
\sref{catalysis} presents results for transferring between activation energies and the classification of the rate determining step.
\sref{discussion} discusses the relative effectiveness of the transfer techniques on these problems, and \sref{conclusion} summarizes the major findings of this work.

\section{Methods} \label{sec:methods}

\begin{figure}[t]
  \input{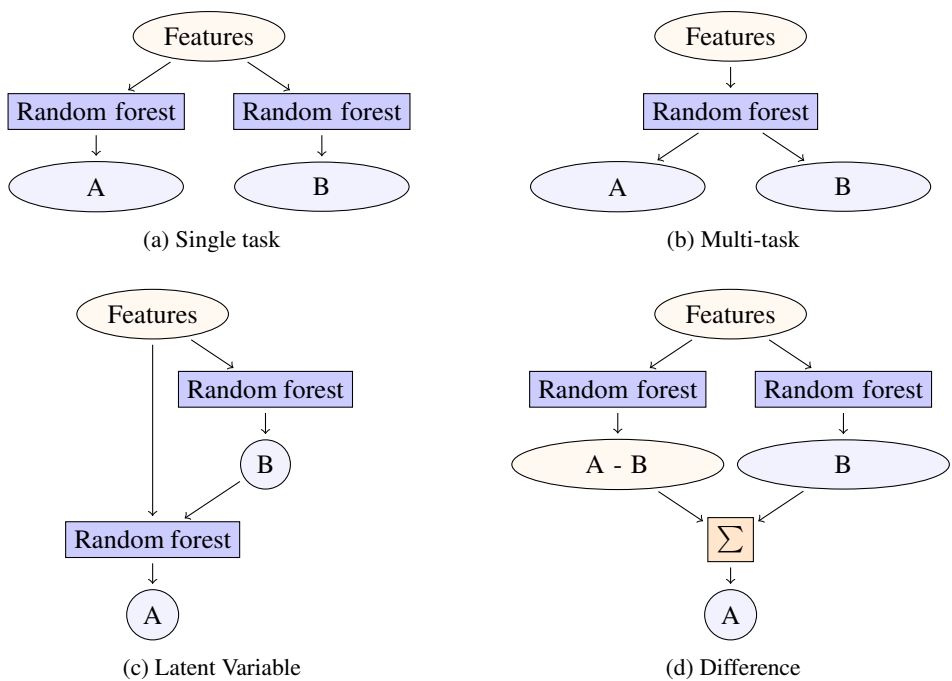}
  \centering
  \begin{subfigure}{0.49\textwidth}
    \centering
    \begin{tikzpicture}[node distance=1cm, auto]
\node[feature] (elem) {\features};
\node[dummy, below of=elem] (dum) {};
\node[ml, right of=dum, node distance=1.5cm, text width=6em] (lolo1) {\lolo};
\node[ml,  left of=dum, node distance=1.5cm, text width=6em] (lolo2) {\lolo};
\node[output, below of=lolo1,text width=4em] (bandgap) {\lv};
\node[output, below of=lolo2,text width=4em] (color) {\resp};

\path (elem) edge[ar] (lolo1);
\path (elem) edge[ar] (lolo2);
\path (lolo1) edge[ar] (bandgap);
\path (lolo2) edge[ar] (color);
\end{tikzpicture}
    \caption{Single task}
  \end{subfigure}
  \begin{subfigure}{0.49\textwidth}
    \centering
    \begin{tikzpicture}[node distance=1cm, auto]
\node[feature] (elem) {\features};
\node[ml, below of=elem] (lolo) {\lolo};
\node[dummy, below of=lolo] (dum) {};
\node[output, right of=dum, node distance=1.5cm, text width=4em] (bandgap) {\lv};
\node[output,  left of=dum, node distance=1.5cm, text width=4em] (color) {\resp};

\path (elem) edge[ar] (lolo);
\path (lolo) edge[ar] (bandgap);
\path (lolo) edge[ar] (color);
\end{tikzpicture}
    \caption{Multi-task}
  \end{subfigure}
  \par \bigskip
  \begin{subfigure}{0.49\textwidth}
    \centering
    \begin{tikzpicture}[node distance=1cm, auto]
\node[feature] (elem) {\features};
\node[dummy, below of=elem] (dum1) {};
\node[dummy, below of=dum1] (dum2) {};
\node[dummy, below of=dum2] (dum3) {};
\node[ml, right of=dum1, node distance=1.5cm,] (lolo1) {\lolo};
\node[output, right of=dum2, node distance=1.5cm] (bandgap) {\lv};
\node[ml, below of=dum2] (lolo2) {\lolo};
\node[output,  below of=lolo2] (color) {\resp};

\path (elem) edge[ar] (lolo1);
\path (elem) edge[ar] (lolo2);
\path (lolo1) edge[ar] (bandgap);
\path (bandgap) edge[ar] (lolo2);
\path (lolo2) edge[ar] (color);
\end{tikzpicture}
    \caption{Latent Variable}
  \end{subfigure}
  \begin{subfigure}{0.49\textwidth}
    \centering
    \begin{tikzpicture}[node distance=1cm, auto]
\node[feature] (elem) {\features};
\node[dummy, below of=elem] (dum) {};
\node[dummy, below of=dum] (dum2) {};
\node[ml, right of=dum, node distance=1.5cm, text width=6em] (lolo1) {\lolo};
\node[ml,  left of=dum, node distance=1.5cm, text width=6em] (lolo2) {\lolo};
\node[output, below of=lolo1,text width=5em] (dft) {\lv};
\node[feature, below of=lolo2,text width=5em] (delta) {\resp~-~\lv};
\node[analytic, below of=dum2] (sum) {$\sum$};
\node[output, below of=sum] (exp) {\resp};

\path (elem) edge[ar] (lolo1);
\path (elem) edge[ar] (lolo2);
\path (lolo1) edge[ar] (dft);
\path (lolo2) edge[ar] (delta);
\path (delta) edge[ar] (sum);
\path (dft) edge[ar] (sum);
\path (sum) edge[ar] (exp);
\end{tikzpicture}
    \caption{Difference}
  \end{subfigure}
  \caption{ \flabel{arch}
    Architectures for three TL techniques and a reference single task architecture.
    In the latent variable and difference architectures, information is transferred from the B task to the A task.
    In the multi-task architecture, it is transferred between the two tasks symmetrically.
  } 
\end{figure}

Transfer learning (TL), also known as inductive transfer, can be used to address sparsity in labeled data by transferring information from one set of labeled data to another~\cite{Pan2010}.
There are many TL techniques that span a variety of learning tasks, \eg pre-training deep neural nets~\cite{pratt1991direct, hinton2006reducing}.
This study focuses on three TL methods that are suitable for the types of regression and classification problems commonly found in materials informatics: explicit latent variables, multi-task learning, and difference learning.

To illustrate these TL techniques, consider two datasets with band gap labels:
\begin{itemize}
  \item \textbf{Materials Project}~\cite{Jain2013}:
Density functional theory (DFT)~\cite{DFT} calculations for 69,640 inorganic materials, of which 49,547 are computed to be stable, performed with VASP~\cite{VASP}.
Records contain structural information in the CIF format and properties, including band gap, total magnetization, bulk and shear modulus, Poisson's ratio, and elastic anisotropy.

  \item \textbf{Strehlow and Cook}~\cite{strehlow1973compilation}: A compilation of 1,449 records from 723 experimental references.
  Some records contain coarse grain structural information, \eg whether the sample was single crystalline or polycrystalline.
  Measurements were taken at a variety of temperatures, with 300K being the majority case.
  A minority of the records (139) include the color of the material as a property.
\end{itemize}

One can expect the DFT band gaps to be correlated with the experimental band gaps, but to have systematic errors.
In that regard, experimental and computational band gaps represent a case of multi-fidelity data.
One can additionally expect the color to be correlated with the band gap, due to its relationship with photo absorption and emission mechanisms.

\paragraph{Explicit latent variables}
Latent variables are features that are predicted by one model, then used as an input to another model (as shown schematically in \fref{arch}).
We distinguish \textit{explicit} latent variables as those for which at least some labels are available in the training data, versus \textit{hidden} latent variables that are inferred as part of the training process, \eg in structural element modeling~\cite{bentler1980linear}, but for which no training labels are available.
In this example, band gap is an explicit latent variable for color.
First, a model is learned for band gap as a function of chemical formula.
Next, missing band gap labels are predicted from that machine learning model.
Finally, a model is learned for color as a function of chemical formula and band gap.
This scheme transfers information from the band gap labels to the color model.

It is also possible to apply the explicit latent variable formalism for multi-fidelity data.
Here, lower fidelity DFT band gaps would be an explicit latent variable for higher fidelity experimental band gaps.
First, a model is learned for DFT band gap as a function of chemical formula.
Next, missing DFT band gap labels are predicted.
Finally, a model is learned for experimental band gap as a function of both chemical formula and DFT band gap.


\paragraph{Multi-task learning}
Multi-task learning is a technique where a single model form is fit to multiple sets of labels~\cite{caruana1998multitask}.
In the example, the models for experimental band gap and computational band gap, or the models for band gap and color would be learned simultaneously and constrained to have similar structure.
Thus, instead of having separate models for each label, there is one model that has multiple responses.
Multi-task learning has been successfully applied with task counts into the hundreds, \eg massively multi-task networks trained on 259 datasets for drug discovery~\cite{ramsundar2015massively}.

Multi-task learning is very common for neural networks and can also be implemented in random forests via multi-task decision trees.
Multi-task trees, like conventional decision trees, are trained by selecting binary partitions of the training data that maximally reduce the impurity of the resulting partitions.
In a multi-task tree, the impurity is simply summed across each of many labels when selecting the optimal split.

\paragraph{Difference learning}
Difference learning is a technique where training data is relabeled by the difference between two labels and a model is learned for the difference.
Difference learning is effective when the difference is a simpler function of the features than the labels themselves.
In the band gap example, one would compute the difference between the computational and experimental band gaps, learn models for the computational band gap and the difference, and then construct a model for the experimental band gap as a sum of the computational band gap and the difference.
Difference learning can be generalized to other invertible operations, \eg the ratio.

Difference learning is a particularly elegant choice for multi-fidelity data, but can be applied to any data of the same type.
Unlike multi-task learning and explicit latent variables, difference learning cannot be used directly for multi-class classification.

\paragraph{Implementations} In each TL architecture, we take the base learners to be a random forest with jackknife-based uncertainty quantification~\cite{Wager2014}.
For multi-task learning, the impurity is expressed as the sum of variances for regression tasks and the sum of Gini impurities for classification tasks, each weighted by the number of labels present for that task.
The learners are implemented in the open source Lolo package~\cite{Lolo}.

Each TL architecture is evaluated with 8-fold cross-validation or a 10\% holdout set.
The folds and holdout set are held constant across the tested architectures and, when applicable, training set sizes.
When transferring from a large model, as will be the case for DFT band gaps, the large model's training data is held fixed to facilitate pre-training.
The selection of folds and holdout sets is repeated in trials to quantify uncertainty.
Error bars are reported at one standard deviation in the estimate of the mean.

\section{Case study: band gap and color of solid state crystalline compounds} \label{sec:bandgap}

\newcommand{\nfold}[0]{8}
\newcommand{\nhold}[0]{37}
\newcommand{\nexp}[0]{374~}
\newcommand{\ndft}[0]{10000~}
\newcommand{\ntrial}[0]{100~}
\newcommand{\ncolor}[0]{60~}

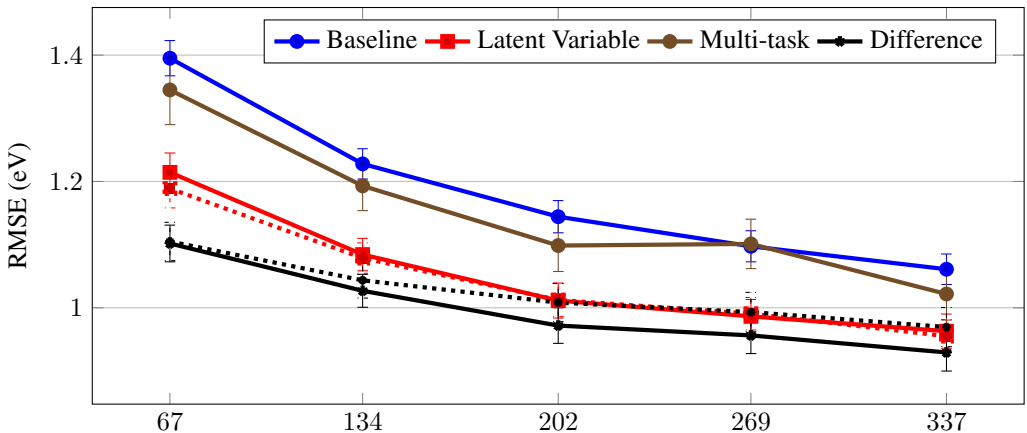
\begin{figure}
  \centering
  \pgfplotstableread[col sep=comma]{data/bandgap-lc.csv}{\lc}
\pgfplotstableread[col sep=comma]{data/bandgap-lc-both.csv}{\lcboth}
\begin{tikzpicture}[scale=1.0]
\begin{axis}[
  xtick=data,
  width=\textwidth,
  height=0.3\textheight,
  legend pos=north east,
  ylabel={\text{RMSE (eV)}}
]
\addplot table [x={Size}, y={Direct}, y error={DirectError}] {\lc};
\addplot table [x={Size}, y={LatentVariable}, y error={LatentVariableError}] {\lc};
\pgfplotsset{cycle list shift=-1}
\addplot plot [dotted] table [x={Size}, y={LatentVariable}, y error={LatentVariableError}] {\lcboth};
\addplot table [x={Size}, y={Multitask}, y error={MultitaskError}]  {\lc};
\addplot table [x={Size}, y={Difference}, y error={DifferenceError}] {\lc};
\pgfplotsset{cycle list shift=-2}
\addplot plot [dotted] table [x={Size}, y={Difference}, y error={DifferenceError}] {\lcboth};
\legend{Baseline, Latent Variable, ,Multi-task, Difference, }
\end{axis}
\end{tikzpicture}
  \caption{\flabel{bandgap-lc}
    RMSE on a 10\% holdout validation set of experimental band gaps as a function of the number of experimental band gap training labels included in the training set.
    The dotted lines indicate the availability of exact DFT labels at predict time.
    The DFT data was sub-sampled to \ndft labels and held constant through the test.
    For the multi-task strategy, results are shown for 15 trials. For all other strategies, results are shown for \ntrial trials.
  }
\end{figure}

A material's band gap is a characteristic of its electronic structure that is particularly important for material selection on the basis of electronic or optical properties.
It is accessible through both experimental measurement and first-principles calculations, \eg DFT.
Due to significant efforts in building databases of high-throughput DFT calculations, such as the Materials Project~\cite{Jain2013}, AFLOW~\cite{curtarolo2012aflowlib}, and OQMD~\cite{saal2013materials,kirklin2015open}, DFT-calculated band gaps are available for a large number of common inorganic compounds.
However, these computational band gaps are known to have significant and systematic deviations from experimentally measured values, limiting their application to band structure design.
Approximations with methods like DFT for other properties have similar limitations.

The data are taken from the two datasets introduced in \sref{methods}~\cite{strehlow1973compilation, DFT}, both accessed from Citrination~\cite{citrination}.
The experimental data were cleaned in three ways.
First, the data was filtered to band gaps and colors measured at 300K, the condition for the majority of measurements in the data.
Next, duplicate measurements of the band gap for the same material and temperature were averaged.
The average variance of the measurements with respect to their material-wise average is about 0.40 eV, which implies a noise level in the experimental labels.
Finally, diversity in color observations was reduced by collecting similar color classes until there were 7 categories with at least 5 instances each.
If multiple color categories were recorded for the same material, those conflicting labels were discarded.
The computational data was filtered to only stable compounds.
After pre-processing, the high-fidelity experimental dataset contained 374 band gap measurements and 60 color observations, and the lower-fidelity computational dataset contained 49,547 calculated band gaps.

First, we evaluate TL architectures for modeling experimental band gaps with added DFT labels.
Because the objective is to improve predictive performance on the experimental band gap values, all validation is performed against only the experimental data.
Our baseline for model performance is a model trained on only the experimental band gaps.
Each random forest is grown to full depth with the number of trees equal to the number of training labels, except for models trained on the DFT band gap dataset, which use 64 trees due to performance considerations.

\fref{bandgap-lc} shows the variation in model performance for the various strategies as a function of experimental dataset size.
For the explicit latent variable and difference architectures, there is a distinction drawn between using model-predicted DFT labels for the experimental materials and having the exact DFT labels at train and predict.
Somewhat counter-intuitively, using only modeled values has similar or better performance in each case.
This highlights the role of the DFT band gap as a consistent descriptor of the material rather than an actual physical property.

Among the TL architectures, the difference architecture has the best performance, followed by the explicit latent variable architecture.
Multi-task learning is considerably less effective, but always at least matches the baseline model performance.
The difference and explicit latent variable architectures are able to replicate baseline performance with between 2$\times$ and 4$\times$ less data.

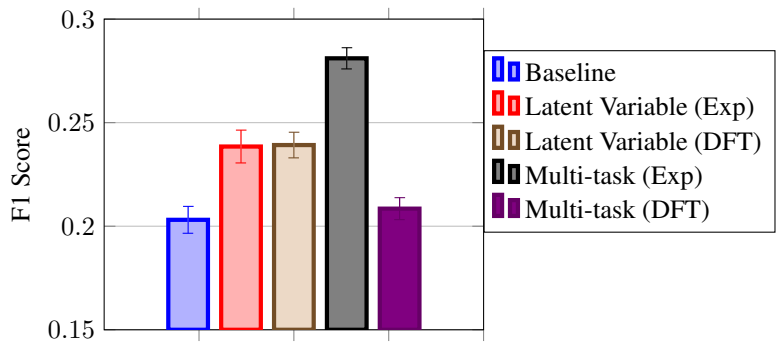
\begin{figure}
  \centering
  \def\baseline{0.31}
\def\baselineError{0.02}

\pgfplotstableread[col sep=comma]{data/color-f1.csv}{\multifidelity}
\begin{tikzpicture}[scale=1.0]
\begin{axis}[
  xtick={},
  xticklabels={},
  height=0.25\textheight,
  legend style={at={(1,0.9)}, anchor=north west},
  legend cell align={left},
  ymin=0.15, ymax=0.3,
  ybar=5pt,
  legend columns=1,
 bar width=15pt,
  enlarge x limits=0.4,
  ylabel={F1 Score}
]

\addplot table [x expr=\coordindex, y={Direct}, y error={DirectError}] {\multifidelity};
\addplot table [x expr=\coordindex, y={LatentVariable}, y error={LatentVariableError}] {\multifidelity};
\addplot table [x expr=\coordindex, y={DFTLatentVariable}, y error={DFTLatentVariableError}] {\multifidelity};
\addplot table [x expr=\coordindex, y={Multitask}, y error={MultitaskError}] {\multifidelity};
\addplot table [x expr=\coordindex, y={DFTMultitask}, y error={DFTMultitaskError}] {\multifidelity};



\legend{Baseline, Latent Variable (Exp), Latent Variable (DFT), Multi-task (Exp), Multi-task (DFT)}
\end{axis}
\end{tikzpicture}
  \caption{\flabel{color-f1}
    Multi-class weighted F1 score of the color labels from 25 trials of \nfold-fold cross-validation on \ncolor experimental training labels.
    Information is transfered from \nexp and \ndft experimental and DFT band gap labels, respectively.
  }
\end{figure}

Additionally, TL is applied to the classification of color, which is sparsely labeled in the experimental dataset, with added band gap labels.
The difference architecture does not apply to classification, so it is skipped.
Both experimental and DFT labels are tested in the latent variable architecture, but, for performance reasons, only experimental labels are used for multi-task.

\fref{color-f1} compares the classification performance for predicting color using the different transfer learning methods.
Multi-task with experimental labels has the strongest performance, but multi-task on DFT labels does not outperform the baseline.  
Explicit latent variables are equally effective with experimental and DFT labels, significantly outperforming the baseline in each case.

The band gap case study is representative of a large class of problems in multi-fidelity modeling, \eg using a large number of inexpensive GGA functional calculations and a modest number of more expensive HSE functional calculations to model the HSE results~\cite{zhu2017fundamental}.
The relative effectiveness of the difference architecture suggests that the DFT calculations used to compute the low fidelity band gaps are missing an additive term that is a simpler function of the material than the band gap itself.

\section{Case study: catalysis of the reduction of nitric oxide} \label{sec:catalysis}

\begin{figure}
  \input{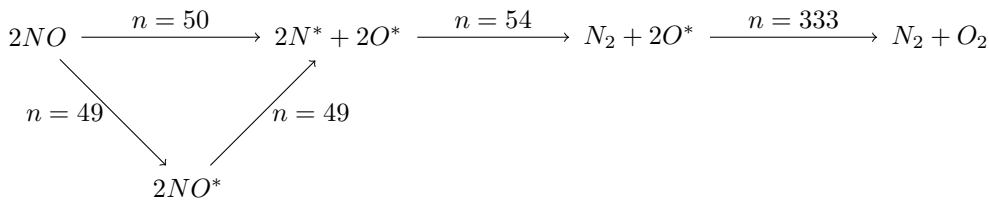}
  \centering
  \begin{tikzpicture}[node distance=4cm, auto]

\node (inputs) {$2NO$};
\node[below of=inputs, node distance=2cm] (dum) {};
\node[right of=dum, node distance=2cm] (half) {$2NO^*$};
\node[right of=inputs, node distance=4cm] (adsorbed) {$2N^* + 2O^*$};
\node[right of=adsorbed] (red1) {$N_2 + 2O^*$};
\node[right of=red1] (final) {$N_2 + O_2$};

\path (inputs) edge[ar] node {$n=50$} (adsorbed);
\path (inputs) edge[ar] node [left] {$n=49$} (half);
\path (half) edge[ar] node [right] {$n=49$} (adsorbed);
\path (adsorbed) edge[ar] node {$n=54$} (red1);
\path (red1)  edge[ar] node {$n=333$} (final);
\end{tikzpicture}
  \caption{ \flabel{reaction}
    Reaction steps in the reduction of $NO$, labeled by the number of catalysts with reaction enthalpies and activation energies in the SUNCAT database.
  }
\end{figure}

Next, we consider the reduction of nitric oxide, NO, to oxygen and nitrogen on a variety of metal surfaces.
Nitric oxide is one component of NOx, a significant contributor to air pollution that is mitigated by catalytic converters.
A better understanding of this catalytic reaction on metal surfaces is therefore of great industrial relevance.

The heterogeneous catalysis community has established that there are linear relationships between a reaction's enthalpy and activation energy, known as Brønsted-Evans-Polanyi (BEP) relations~\cite{Logadottir2001,Bligaard2004}.
There are also linear relationships between the activation energies of different reactions, known as scaling relations, that are consistent at different material surfaces~\cite{Greeley2016,Norskov2004}.
These relations reinforce the notion that transfer should be possible between regression tasks for enthalpies and activation energies on different reactions.
However, the universality of the relation with respect to the catalyst, \ie that the ratio does not depend on the material, could limit the ability of the tasks to identify different representations of the input space.
Therefore, we expect benefit most from TL when the labels for different tasks correspond to different materials, maximizing the coverage of the input space over all of the tasks.

Density functional theory (DFT) calculations of individual reaction steps inform mechanistic understanding of this reaction and guide future catalyst design.
In recent years, the SUNCAT Center for Interface Science and Catalysis has built a database for DFT calculations of reaction step enthalpies and activation energies on metal surfaces~\cite{hummelshoj2012catapp}.
This database has over 4000 calculations and is available on Citrination~\cite{citrination}.

\begin{figure}
  \input{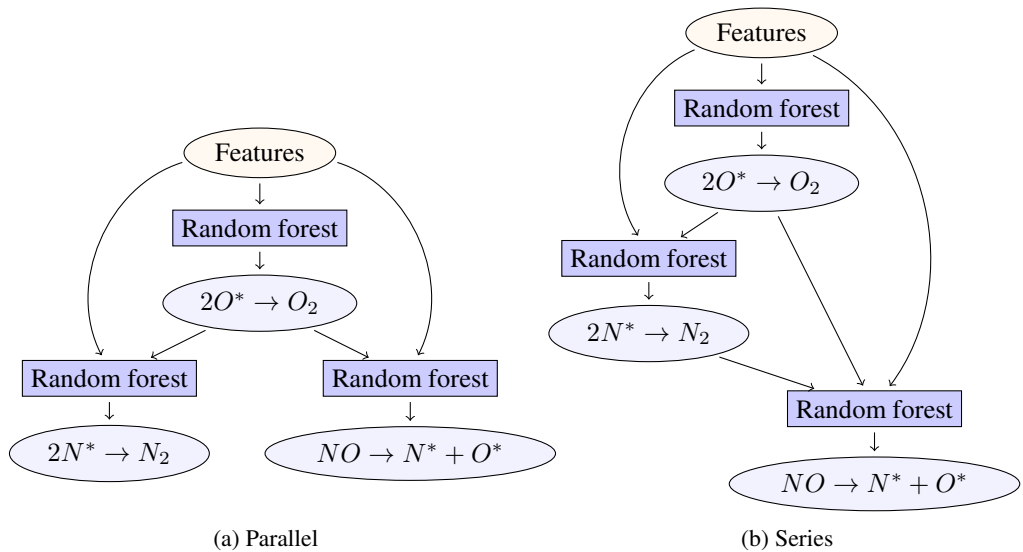}
  \centering
  \begin{subfigure}[b]{0.49\textwidth}
    \centering
    \begin{tikzpicture}[node distance=1cm, auto]
\node[feature] (elem) {\features};
\node[ml, below of=elem] (lolo1) {\lolo};
\node[output, below of=lolo1] (oxygen) {$2O^* \rightarrow O_2$};
\node[dummy, below of=oxygen] (dum1) {};
\node[ml, left of=dum1, node distance=2.0cm,] (lolo2) {\lolo};
\node[output, below of=lolo2] (nitrogen) {$2N^* \rightarrow N_2$};
\node[ml, right of=dum1, node distance=2.0cm,] (lolo3) {\lolo};
\node[output, below of=lolo3] (nox) {$NO \rightarrow N^* + O^*$};

\path (elem) edge[ar] (lolo1);
\path (elem) edge[ar, bend right=50] (lolo2);
\path (elem) edge[ar, bend left=50] (lolo3);
\path (lolo1) edge[ar] (oxygen);
\path (oxygen) edge[ar] (lolo2);
\path (oxygen) edge[ar] (lolo3);
\path (lolo2) edge[ar] (nitrogen);
\path (lolo3) edge[ar] (nox);
\end{tikzpicture}
    \caption{Parallel}
  \end{subfigure}
  \begin{subfigure}[b]{0.49\textwidth}
    \centering
    \begin{tikzpicture}[node distance=1cm, auto]
\node[feature] (elem) {\features};
\node[ml, below of=elem] (lolo1) {\lolo};
\node[output, below of=lolo1] (oxygen) {$2O^* \rightarrow O_2$};
\node[dummy, below of=oxygen] (dum1) {};
\node[dummy, below of=dum1] (dum2) {};
\node[dummy, below of=dum2] (dum3) {};
\node[ml, left of=dum1, node distance=1.5cm,] (lolo2) {\lolo};
\node[output, below of=lolo2] (nitrogen) {$2N^* \rightarrow N_2$};
\node[ml, right of=dum3, node distance=1.5cm,] (lolo3) {\lolo};
\node[output, below of=lolo3] (nox) {$NO \rightarrow N^* + O^*$};

\path (elem) edge[ar] (lolo1);
\path (elem) edge[ar, bend right=50] (lolo2);
\path (elem) edge[ar, bend left=50] (lolo3);
\path (lolo1) edge[ar] (oxygen);
\path (oxygen) edge[ar] (lolo2);
\path (oxygen) edge[ar] (lolo3);
\path (lolo2) edge[ar] (nitrogen);
\path (nitrogen) edge[ar] (lolo3);
\path (lolo3) edge[ar] (nox);
\end{tikzpicture}
    \caption{Series}
  \end{subfigure}
  \caption{ \flabel{catapp-arch}
    Two latent variable architectures for activation energies of reaction steps.
    In the parallel case, the $2O^* \rightarrow O_2$ step is chosen as the latent variable due to its relatively plentiful training data.
    In the series case, the ordering transfers information in order of decreasing label count.
  }
\end{figure}

There are five relevant reaction steps present in the SUNCAT database, which are shown in \fref{reaction}.
Between the two pathways, the $NO \rightarrow N^* + O^*$ reaction is generally preferred, so we focus on the three reactions on the upper pathway.
The $2O^* \rightarrow O_2$ step, which is relevant in other applications, \eg fuel cells, has significantly more training data, so our goal is to transfer information from it to the other more scarcely labeled tasks.
Labels for the $2N^* \rightarrow N_2$ and $NO \rightarrow N^* + O^*$ reaction steps cover similar catalysts, so we do not expect to benefit from information transfer between them.

The approximate symmetry in the number of labels for the $2N^* \rightarrow N_2$ and $NO \rightarrow N^* + O^*$ reactions suggests two possible latent variable architectures: a parallel architecture where the oxygen step alone is intermediate for both other steps and a series architecture transferring from the great to the least labeled training tasks, as shown in \fref{catapp-arch}.

\begin{figure}
  \begin{subfigure}{0.63\textwidth}
    \pgfplotstableread[col sep=comma]{data/catapp-rmse.csv}{\corr}
\begin{tikzpicture}[scale=1.0]
\begin{axis}[
  xtick=data,
  width=\textwidth,
  height=0.3\textheight,
  xticklabels={$2N^* \rightarrow N_2$, $NO \rightarrow N^* + O^*$},
  ybar=5pt,
  legend pos=north west,
  ymax=1.0,
  enlarge x limits=0.5,
  restrict x to domain=0:1,
  legend columns=3,
  ylabel={RMSE (eV)}
]

\addplot table [x expr=\coordindex, y={Direct}, y error={Direct Error}] {\corr};
\addplot table [x expr=\coordindex, y={Stacked}, y error={Stacked Error}] {\corr};
\addplot table [x expr=\coordindex, y={Series}, y error={Series Error}] {\corr};
\addplot table [x expr=\coordindex, y={Multitask}, y error={Multitask Error}] {\corr};
\addplot table [x expr=\coordindex, y={Difference}, y error={Difference Error}] {\corr};

\legend{Baseline, Parallel, Series, Multi-task, Difference}
\end{axis}
\end{tikzpicture}
  \end{subfigure}
  \begin{subfigure}{0.36\textwidth}
    \pgfplotstableread[col sep=comma]{data/catapp-rmse.csv}{\corr}
\begin{tikzpicture}[scale=1.0]
\begin{axis}[
  xtick={4},
  width=\textwidth,
  height=0.3\textheight,
  xticklabels={RDS},
  ybar=5pt,
  ylabel near ticks,
  yticklabel pos=right,
  enlarge x limits=0.2,
  restrict x to domain=4:4,
  ylabel={F1 score}
]

\addplot table [x expr=\coordindex, y={Direct}, y error={Direct Error}] {\corr};
\addplot table [x expr=\coordindex, y={Stacked}, y error={Stacked Error}] {\corr};
\addplot table [x expr=\coordindex, y={Series}, y error={Series Error}] {\corr};
\addplot table [x expr=\coordindex, y={Multitask}, y error={Multitask Error}] {\corr};
\addplot table [x expr=\coordindex, y={Difference}, y error={Difference Error}] {\corr};

\end{axis}
\end{tikzpicture}
  \end{subfigure}
  \caption{ \flabel{catapp-res}
    RMSE of regression models for reaction step activation energies and multi-class F1 score for the composite task of identifying the rate determining step (RDS) for each of four TL techniques.
    Error bars are computed by repeating the procedure 7 times.
  }
\end{figure}
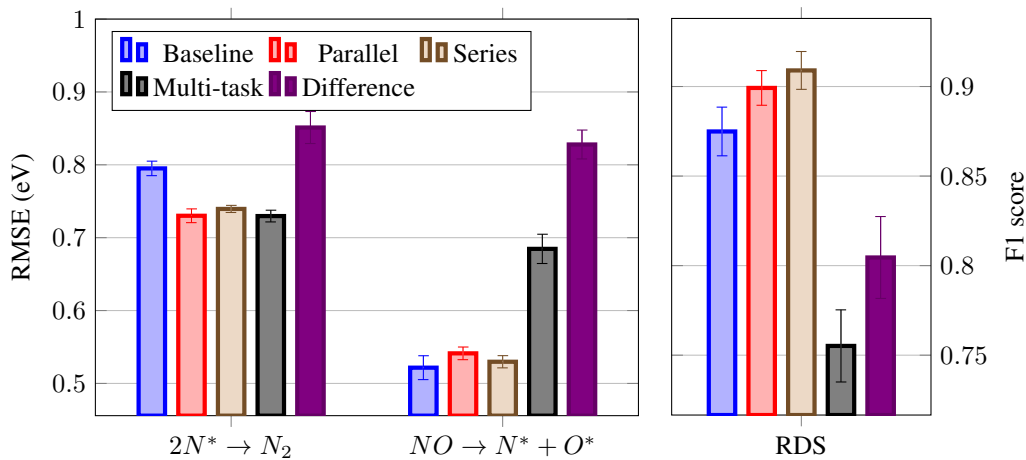

A regression model is trained for the activation energy of each step following each TL architecture.
When the database includes activation energies from multiple binding sites, the lowest activation energy is chosen.
Each random forest is grown to full depth with two trees per training label.
The $2O^* \rightarrow O_2$ model is identical in the direct, latent variable, and difference architectures and does not change significantly in the multi-task case due to the label imbalance.
The model performance for the other two activation energies depends on the architecture, favoring the two latent variable and multi-task for the $N^* \rightarrow N$ step and the direct and series architectures in the $NO \rightarrow N^* + O^*$ step, as shown in \fref{catapp-res}.
The difference architecture, which had the best performance in the band gap case study, had significantly worse performance here than the baseline model.

In addition to learning regression models for the activation energy, the rate determining step (RDS) is identified by selecting the reaction step with the lowest predicted activation energy.
If the regression tasks have correlated outputs, then the RDS classification task could enjoy a cancellation of errors.
Indeed, the two latent variable architectures, in which the activation energies are coupled explicitly, have significantly better F1 scores than the reference non-TL architecture.
The multi-task architecture, despite being more accurate than the difference architecture for each individual reaction, performs significantly worse than the other architectures on the RDS classification task.
Either the difference architecture is outperforming on RDS classification due to cancellation of error, or the multi-task architecture is underperforming.


The reaction energetics case study is representative of a large class of correlated tasks.
While the difference architecture did not perform well, a generalization of the difference to be with respect to the predictions of a linear model trained between the reaction rates could better capture perturbations around the expected linear scaling.

\section{Discussion} \label{sec:discussion}

The learning curves in \fref{bandgap-lc} demonstrate that transfer learning can be effective with tens to hundreds of training instances.
The relative improvement compared to the non-transfer architecture decreases with high fidelity label size, but only weakly.
This is consistent with studies demonstrating the effectiveness of transfer learning on larger datasets~\cite{montavon2013machine, ramsundar2015massively}, which were not the subject of this study.

The ability to transfer information to very small sets of training labels could be particularly valuable to experimental design processes, where machine learning has been shown to improve efficiency by 2-4x when there are only tens of labels~\cite{ling2017high}.
Experimental design is most effective when uncertainty estimates are available, which motivates additional study of uncertainty quantification in these transfer learning methods.

Multi-task learning can be considerably more expensive than explicit latent variables and difference learning.
In the case of transferring from a plentiful label to a scarce label, explicit latent variable and difference architectures can re-use a pre-trained model on the plentiful label.
In contrast, multi-task learning requires the full multi-task model to be retrained for each unique set of scarce labels.
This is particularly relevant during cross-validation of the scarce label, where the multi-task model must be retrained on each fold.
Similarly, sequential learning on a small but growing number of scarce labels requires many multi-task retrains.
Preference might be given to explicit latent variable or difference architectures when the cost is a consideration and multiple trainings are anticipated.

The relatively poor performance of the multi-task architecture when transferring information from DFT labels may be due to the large label count imbalance between the experimental and DFT data.
The multi-task method allows one task to take priority over another.
If the tasks are not sufficiently correlated in the neighborhood of their labels, performance can degrade.
To mitigate this, the two classes of labels could be re-weighted by a constant factor to reduce or eliminate the imbalance, creating a hyperparameter.
It seems likely that the optimal value of this hyperparameter is not the trivial one used here implicitly.
However, the addition of a hyperparameter can add significantly the cost of training a model, which is compounded by the inability to pre-train multi-task decision trees.

Both latent variable and difference architectures provide additional interpretability over single and multi-task learning.
The feature importance of and, in the case of differentiable models, gradient with respect to latent variables is computable, directly quantifying the strength of the relationship between labels.
Difference and ratio architectures learn models for relationships between the labels directly.
For example, the difference model between fidelities can be used to identify regions of input space where the low fidelity approximations break down.
Multi-task is not as readily interpretable, but its effectiveness compared to single task can indicate the degree to which two labels depend on similar representations of the input space.

The explicit relationships between labels in latent variable and difference architectures can couple predictions.
For functions defined on multiple predicted labels, \eg the classification of the rate determining step, the coupling can lead to cancellation of errors.
This behavior is seen in the performance of the explicit latent variable architectures.
Interestingly, however, the multi-task architecture does not appear to benefit from this type of cancellation, at least not to the degree that the difference architecture does.

\section{Conclusions} \label{sec:conclusion}

Transfer learning techniques expand the impact of published data by transferring information between even entirely disjoint datasets.
For multi-fidelity modeling of experimental band baps, the difference architecture achieved comparable performance to a single task model with 4$\times$ fewer experimental labels.
For models of reaction step activation energies for nitric oxide reduction, the latent variable architecture reduces error averaged over the steps by 5\% and improves F1 score of classifying the rate determining step from 0.87 to 0.92.
Given the high cost of experiments in material science applications, it is particularly valuable that transfer learning leverages pre-existing and lower fidelity data sources.

The TL architectures studied here are each more and less effective for different problems.
The difference architecture is the most effective for multi-fidelity band gaps but worse than baseline for activation energies.
The multi-task architecture is the most effective for transferring band gaps to sparse color labels, but worse than baseline for identifying the rate determining reaction step.
The latent variable architecture is the most effective for correlated reaction step activation energies and consistently outperforms the baseline non-transfer architecture.

The explicit latent variable and difference architectures are two examples of a broader class of model architectures that represent relationships between features and labels as a directed graph.
The difference architecture can be generalized as computing additive or multiplicative corrections to the output of any other model, \eg an empirical model.
Likewise, the relationship between the low fidelity and target labels can itself assume any form. If it is chosen to be a machine learned model, then the explicit latent variable formalism is recovered.
In this way, complex relationships between labels from disjoint data sources can be modeled with a combination of low bias, \ie machine learned, and high bias, \eg empirical, functional forms.
In addition to performance gains from transfer learning, these general architectures are well suited for capturing domain knowledge and facilitating model interpretation, both critical capabilities in the low data limit.

While it has been demonstrated that transfer learning as a class of techniques is effective at strengthening models trained on sparse labels, the band gap and catalysis case studies are insufficiently generalizable to support strong guidance for which technique to use in a particular circumstance.
The choice of transfer learning technique is thus an architectural hyperparameter and should ideally be systematically varied to identify the most predictive model.
In the absence of systematic selection, the explicit latent variable architecture offers robust improvement over the single task baseline.
Further development of transfer learning techniques could improve generalizable performance, \eg improving latent variable performance on multi-fidelity data to reach parity with the difference architecture.
Further application of transfer learning to a variety of problems in materials informatics would improve the community's ability to select the most appropriate technique.

\bibliographystyle{apsrev}

{\small
\bibliography{biblio}
}

\end{document}